\documentclass[letterpaper, 10 pt, conference]{ieeeconf}  

\IEEEoverridecommandlockouts                              

\usepackage{graphics} 
\usepackage{epsfig} 
\usepackage{mathptmx} 
\usepackage{times} 
\usepackage{amsmath} 
\usepackage{amssymb}  
\usepackage{multirow}
\usepackage{booktabs}

\usepackage{etoolbox}

\AtBeginEnvironment{equation}{\small}
\AtBeginEnvironment{align}{\small}

\title{\LARGE \bf
CETUS: Causal Event-Driven Temporal Modeling With Unified Variable-Rate Scheduling
}

\author{
Hanfang Liang\textsuperscript{1},
Bing wang\textsuperscript{1},
Shizhen Zhang\textsuperscript{1},
Wen Jiang\textsuperscript{2},\\
Yizhuo Yang, \textsuperscript{3}
Weixiang Guo,\textsuperscript{3}
Shenghai Yuan\textsuperscript{3,*}, \textit{Member} IEEE
\thanks{\textsuperscript{1}Jianghan University, Wuhan, China.}
\thanks{\textsuperscript{2}Beijing Institute of Technology, China.}
\thanks{\textsuperscript{3}Nanyang Technological University, Singapore.}
\thanks{*Corresponding author: Shenghai Yuan (email: shyuan@ntu.edu.sg).}
}

\begin{document}

\maketitle
\thispagestyle{empty}
\pagestyle{empty}

\begin{abstract}

Event cameras capture asynchronous pixel-level brightness changes with microsecond temporal resolution, offering unique advantages for high-speed vision tasks. 
Existing methods often convert event streams into intermediate representations such as frames, voxel grids, or point clouds, which inevitably require predefined time windows and thus introduce window latency. 
Meanwhile, pointwise detection methods face computational challenges that prevent real-time efficiency due to their high computational cost. 
To overcome these limitations, we propose the Variable-Rate Spatial Event Mamba, a novel architecture that directly processes raw event streams without intermediate representations. 
Our method introduces a lightweight causal spatial neighborhood encoder to efficiently capture local geometric relations, followed by Mamba-based state space models for scalable temporal modeling with linear complexity. 
During inference, a controller adaptively adjusts the processing speed according to the event rate, achieving an optimal balance between window latency and inference latency.

\begin{keywords}
Event camera, state space models, real-time segmentation, direct processing, Mamba architecture
\end{keywords}

\end{abstract}

\section{INTRODUCTION}
Event cameras have emerged as a revolutionary sensing modality for capturing visual information in challenging scenarios where conventional frame-based cameras fail {\cite{huang2023anti}, \cite{jiang2101anti}}. These bio-inspired sensors respond asynchronously to logarithmic brightness changes at each pixel, offering microsecond temporal resolution, high dynamic range exceeding 120 dB, and extremely low power consumption {\cite{wang2025object, gallego2020event, posch2010qvga}}. These characteristics make event cameras particularly suitable for tasks demanding real-time performance such as UAV target detection, and have attracted widespread attention in autonomous driving, robotics, and surveillance applications {\cite{lu2024flexevent}}. However, efficiently utilizing event data for target detection remains a challenging problem. Event streams lack the pixel intensity information of traditional images and differ in data format from conventional frame sequences, which prevents the direct application of existing vision detection algorithms.

\begin{figure}[!t]
\centering
\includegraphics[width=3.4in]{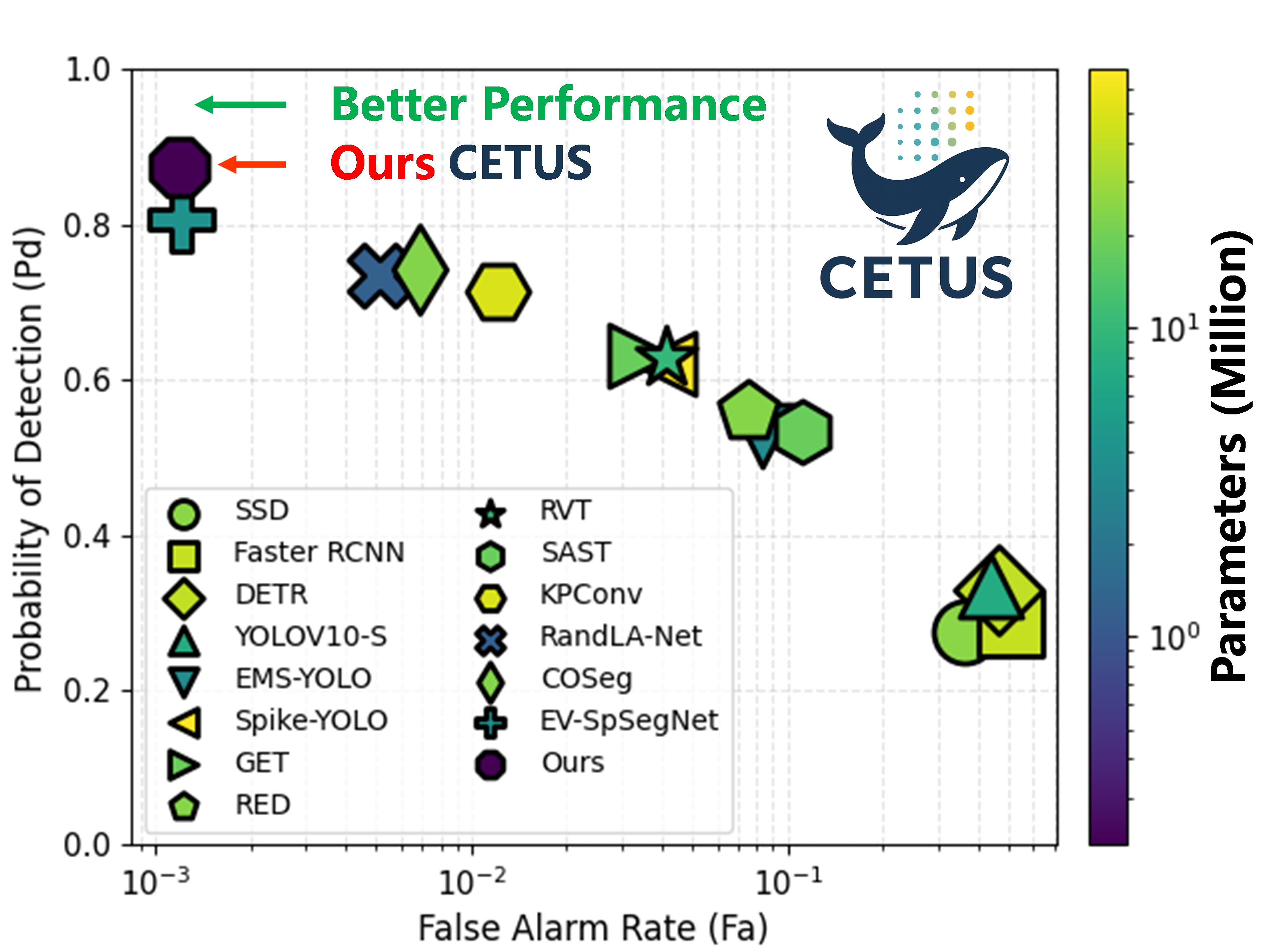}
\vspace{-20pt}
\caption{
Our lightweight model achieves higher detection probability and a lower false alarm rate with significantly reduced number of parameters, while traditional frame-based architectures usually incur greater computational cost.
}
\label{PdvsFa}
\vspace{-15pt}
\end{figure}

Most \textbf{existing} event camera detection methods adapt event data to conventional deep learning architectures by converting the asynchronous stream into synchronous dense representations, such as accumulating events within fixed time windows to generate event frames {\cite{gehrig2023recurrent, liang2025label, perot2020learning, peng2023get, luo2024integer, peng2024scene, su2023deep}}, point clouds or voxel grids \cite{hu2020randla, thomas2019kpconv}. But the fixed time window accumulation strategy artificially reduces the temporal resolution of event cameras by synchronizing context to low frame rates, inevitably neglecting the rich temporal details contained in high-frequency events. Second, window accumulation introduces additional algorithmic latency—for instance, commonly used 50\,ms time windows imply at least 50\,ms of perception lag \cite{lu2024flexevent}, which significantly reduces system response speed for applications such as high-speed UAV capture. For UAVs flying at speeds ranging from 40\,km/h to 80\,km/h, even a window latency of 50\,ms can lead to a localization error of approximately 1m. 
Voxel-based representations discretize events into 3D spatiotemporal grids, maintaining some temporal information but suffering from memory inefficiency and quantization artifacts. Point clouds approaches treat events as unordered sets of 3D points, better preserving the data structure but lacking explicit temporal modeling and struggling with the massive data volumes generated by modern event cameras \cite{an2024rethinking, liang25ICASSP}. 

The \textbf{key challenges} lie in the fact that conversion-based paradigms fundamentally contradict the asynchronous, continuous nature of event data and introduce unnecessary computational overhead and latency. Moreover, these approaches present an inherent trade-off: smaller windows preserve temporal detail but yield sparse, noisy frames, while larger windows produce denser representations at the cost of motion blur and temporal ambiguity. Both choices sacrifice the microsecond temporal resolution that makes event cameras valuable for real-time perception.




These fundamental limitations motivate our approach to develop a window-free detection framework that preserves the native advantages of event cameras. We propose CETUS (Causal Event-driven Temporal modeling with Unified variable-rate Scheduling), a novel event-based object detection framework. 

Instead of converting events into intermediate representations, our method directly processes the asynchronous event stream. We explicitly incorporate the event rate as a temporal feature to better adapt to the high-dynamic output of event cameras. A lightweight neighborhood encoder is designed to capture spatial information by encoding only local neighbor features, thereby avoiding large-window accumulation, reducing latency, and lowering computational overhead. Furthermore, we leverage the Mamba state-space model to combine the benefits of parallel and sequential inference, enabling the network to adapt its inference speed according to the input event rate. This variable-rate inference mechanism balances window latency and inference latency. Experimental results demonstrate that our approach achieves state-of-the-art accuracy and efficiency on UAV detection tasks, as shown in the Fig.~\ref{PdvsFa}.

The main contributions of this paper are as follows:

\begin{itemize}
  \item \textbf{Low-latency event processing:} We eliminate event accumulation or voxelization, directly operating on raw asynchronous streams without fixed windows, thereby avoiding $>$50\,ms latency and achieving millisecond-level detection response.  

 \item \textbf{Event-rate aware neighborhood encoder:} We incorporate event rate as an explicit feature and design a causal lightweight encoder to capture local spatio-temporal relations with minimal computational overhead.

  \item \textbf{Variable-rate inference controller:} We develop a controller that adaptively balances sampling latency and inference latency, enabling stable detection across diverse scenarios. Eliminating the window latency inherent in existing approaches.  

  \item \textbf{Open-source commitment:} We will release our code to foster community development in real-time event-based perception.

\end{itemize}

\section{Related Works}
Event-based vision algorithms can be categorized into two fundamental paradigms based on their data processing strategies: methods that preserve the sparse nature of events and those that convert events into dense representations. We review both categories and position our approach within this landscape. 

\subsection{Event representations}
A common practice in event-based object detection is to convert the asynchronous stream into dense tensors (event frames/voxel grids) at fixed time windows and then apply CNN/Transformer-style backbones. Recurrent Vision Transformers (RVT) exemplify this line: they process windowed event representations with recurrent temporal aggregation to cut latency while preserving accuracy compared to earlier dense pipelines {\cite{gehrig2023recurrent}}. More recent work addresses the computational burden of dense attention by sparsifying tokens. SAST (Scene-Adaptive Sparse Transformer) proposes window–token co-sparsification and masked sparse self-attention to keep only informative regions, improving the accuracy–efficiency trade-off on large-scale event detection benchmarks {\cite{peng2024scene}}. Despite these gains, windowed aggregation remains prevalent, which inherently introduces algorithmic latency and may smear high-frequency temporal cues under fast motion.

\subsection{Sparse/SSM backbones for event streams}
State-space models (SSMs) and sparse attention have been explored to capitalize on event sparsity. SMamba integrates Mamba blocks with a Spatio-Temporal Continuity Assessment (STCA) to drop uninformative tokens and prioritizes local scans among high-information tokens, achieving strong accuracy–efficiency balance on Gen1/1Mpx/eTram while reducing overhead versus dense Transformers {\cite{yang2025smamba}}. These advances highlight the promise of SSM-style sequence modeling for events; yet most reported pipelines still operate on batched/windowed tensors rather than strictly stream-causal point processing.

\subsection{Spiking neural networks for low-latency detection}
SNN detectors are attractive for event data due to event-driven computation. Early work (Spiking-YOLO) demonstrated competitive detection with energy benefits, introducing training stabilizers tailored to spikes {\cite{kim2020spiking}}. Follow-ups push end-to-end trainable SNN detectors and hybrid conversions for higher accuracy under low time steps {\cite{yuan2024trainable}}. EAS-SNN further reframes adaptive event sampling as a learnable temporal mechanism inside recurrent convolutional SNNs, and showing transferability to non-spiking backbones {\cite{wang2024eas}}. While SNNs reduce latency/energy, their absolute accuracy often lags the best ANN/Transformer/SSM detectors, and training stability can be challenging at scale.

\subsection{Multimodal event–RGB and frequency-adaptive detection}
Combining frame semantics with event dynamics is another active thread. FlexEvent targets arbitrary-frequency detection via FlexFuser (adaptive event-frame fusion) and FAL (frequency-adaptive learning), maintaining accuracy from 20–90\,Hz and remaining reliable up to 180\,Hz where fixed-frequency baselines (e.g., RVT) degrade {\cite{lu2024flexevent}}. HDI-Former is a hybrid ANN–SNN Transformer that incorporates cross-modal interaction to enable energy-efficient frame–event detection, reporting gains over 11 prior methods on DSEC-Detection {\cite{li2024hdi}}. YOLO-style adaptations also appear, e.g., Recurrent-YOLOv8 that injects recurrency to capture event temporal context on top of a strong frame baseline {\cite{silva2025recurrent}}. These systems demonstrate the value of multimodality and rate-adaptation, but typically retain windowed batching and heavier fusion modules that complicate ultra-low-latency deployment on small aerial platforms \cite{liu2025eventgpt}.

\subsection{Discussion and positioning}
For high-speed UAV detection, three key challenges remain prominent:  

(1) \textbf{Fixed windowing.} Existing frame- or voxel-based methods rely on predefined time windows, which inevitably introduce tens of milliseconds of algorithmic latency and dilute the intrinsic microsecond temporal resolution of event cameras.  

(2) \textbf{Rate variability.} Real-world event streams exhibit large fluctuations in event rate, ranging from sparse low-activity scenarios to dense high-dynamic motion, yet most existing pipelines cannot adapt their inference speed to such variability in a strictly causal manner.  

(3) \textbf{Lightweight modeling.} Real-time deployment requires lightweight architectures with localized spatio-temporal encoders that capture geometric structure efficiently, avoiding the overhead of global attention or dense 3D voxelization. 

Our framework is designed around these challenges: a window-free causal neighborhood encoder preserves fine temporal detail, while an event-rate–aware Mamba backbone provides efficient temporal modeling. Unlike prior detectors that rely on fixed slices or heavy fusion to stabilize time, our approach directly exploits the high temporal fidelity of event cameras with adaptive-speed inference, achieving a favorable balance of accuracy, latency, and efficiency.

\section{Proposed Method}

\begin{figure*}[!t]
\centering
\includegraphics[width=7in]{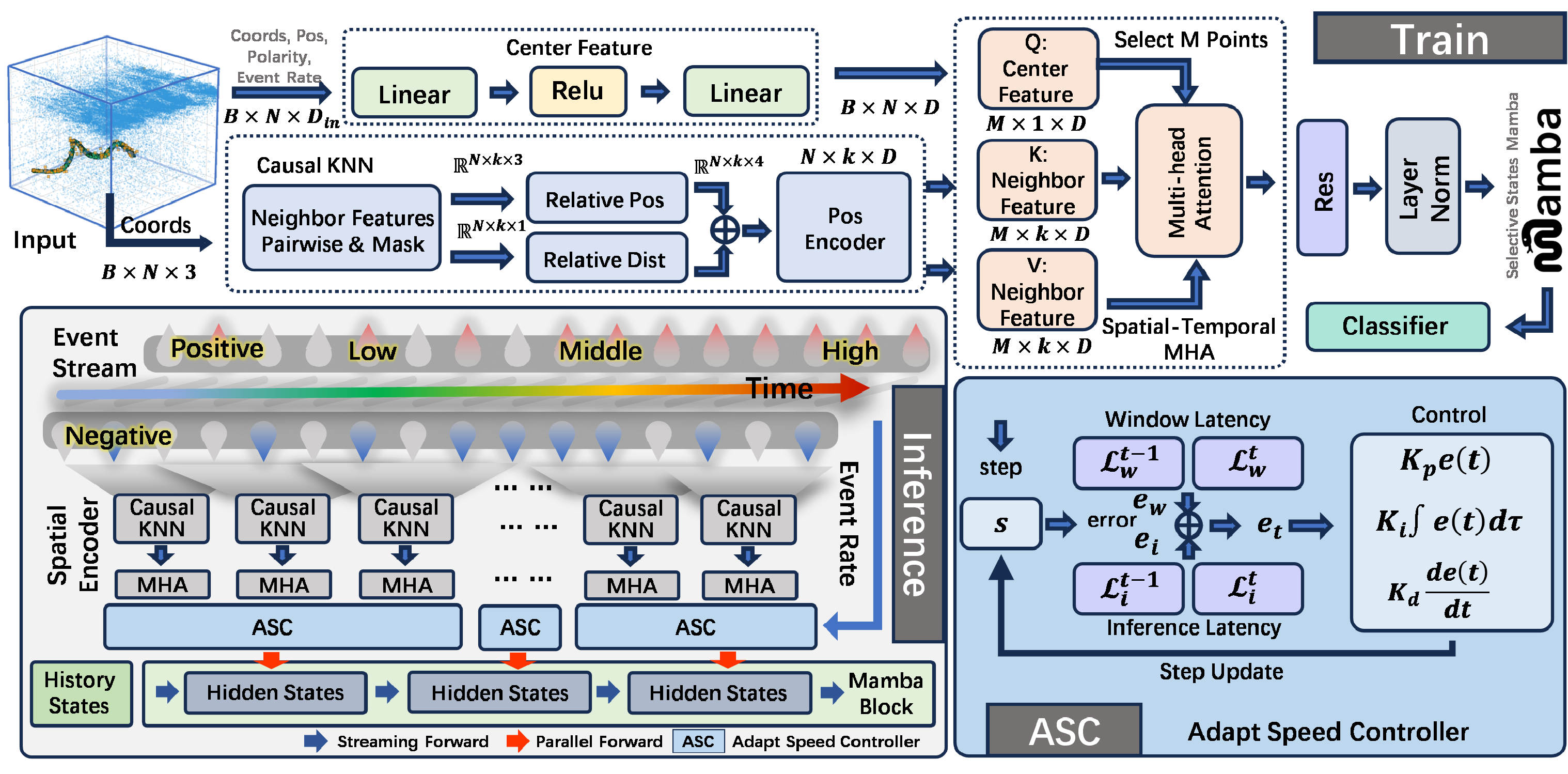}
\vspace{-23pt}
\caption{Overall architecture of Variable-Rate Spatial Event Mamba with three main components: Spatial Encoder, Mamba-based Temporal Modeling, and Adaptive Speed Controller. (i) \textbf{Spatial Encoder}, which encodes raw event streams through causal KNN (K-Nearest Neighbor) and positional encoding to obtain localized spatial features; 
(ii) \textbf{Mamba-based Temporal Modeling}, which stacks state-space blocks with residual and normalization layers to capture long-range temporal dependencies, followed by a lightweight classifier head for detection; 
(iii) \textbf{Inference with Adapt Speed Controller (ASC)}, which regulates the progression step size by jointly monitoring window and inference latency via a PID formulation, and co-adapts the KNN history size, thereby balancing latency and throughput under highly variable event rates.
}
\vspace{-15pt}
\label{2}
\end{figure*}

In our approach, we propose a window-free and variable-rate event detection framework that directly processes asynchronous event streams, illustrated in Fig.~\ref{2}. The method effectively balances inference latency and window latency, achieving state-of-the-art performance using only a minimal number of parameters. We first introduce a lightweight spatial encoder that performs pointwise local propagation to efficiently capture spatial context from raw event streams, making it well-suited to the highly temporal resolution of event cameras. The encoded features are then fed into a state-space model (Mamba) to perform long-range temporal modeling. 

\begin{figure}[!t]
\centering
\includegraphics[width=3.6in]{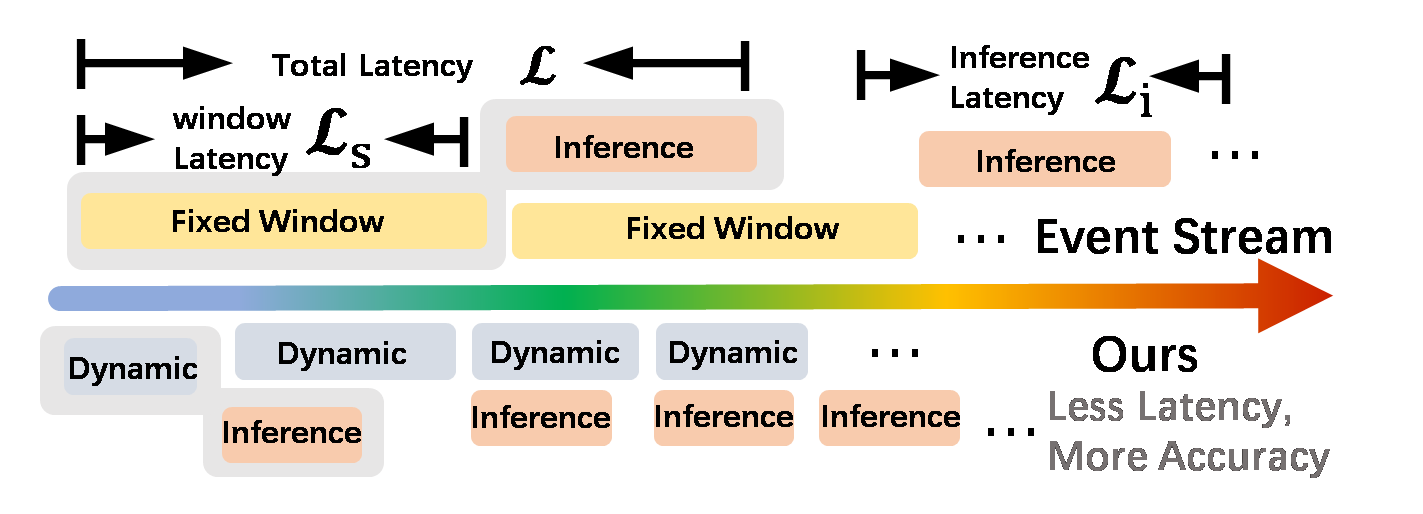}
\vspace{-25pt}
\caption{Fixed-window methods incur large window latency, while our variable-rate inference dynamically adapts to the event stream, reducing delay and maintaining real-time performance.
}
\label{latency}
\vspace{-15pt}
\end{figure}

During inference, since state-space networks support both parallel and sequential propagation, we design a Parallel Group Controller to regulate the scale of pointwise progression. This allows the framework to achieve lower inference latency under reduced window delay, without sacrificing detection performance.

\subsection{Problem Definition}

Let \(\mathcal{E}=\{e_i\}_{i=1}^{N}\) denote an asynchronous event stream, where each event
\(e_i=(x_i,y_i,t_i,p_i)\) encodes polarity \(p_i\) change at pixel \((x_i,y_i)\) with timestamp \(t_i\).  
We define the input feature sequence as \(\mathbf{F}\in\mathbb{R}^{B\times N\times D_{\text{in}}}\), where  
\(B\) is the batch size, \(N\) is the number of events in a chunk, and \(D_{\text{in}}\) is the input feature dimension.  
The corresponding coordinates are represented as \(\mathbf{C}\in\mathbb{R}^{B\times N\times 3}\).  
In addition, we derive an event-rate feature denoted as \(\mathcal{R}\).

We decompose the overall latency into event latency \(\mathcal{L}_{e}\), sampling window latency \(\mathcal{L}_{s}\), and inference speed latency \(\mathcal{L}_{i}\), as shown in Fig.~\ref{latency}. Since \(\mathcal{L}_{e}\) is typically below 1\,ms, it can be ignored. The window latency \(\mathcal{L}_{s}\) is determined by the modeling strategy (around 50\,ms for fixed windows), while the inference latency \(\mathcal{L}_{i}\) depends on the network design and parameter size. Overall, the total latency is given by \(\mathcal{L} = \mathcal{L}_{\text{e}} + \mathcal{L}_{\text{s}} + \mathcal{L}_{\text{i}}\).

\begin{figure}[!t]
\centering
\includegraphics[width=2.0in]{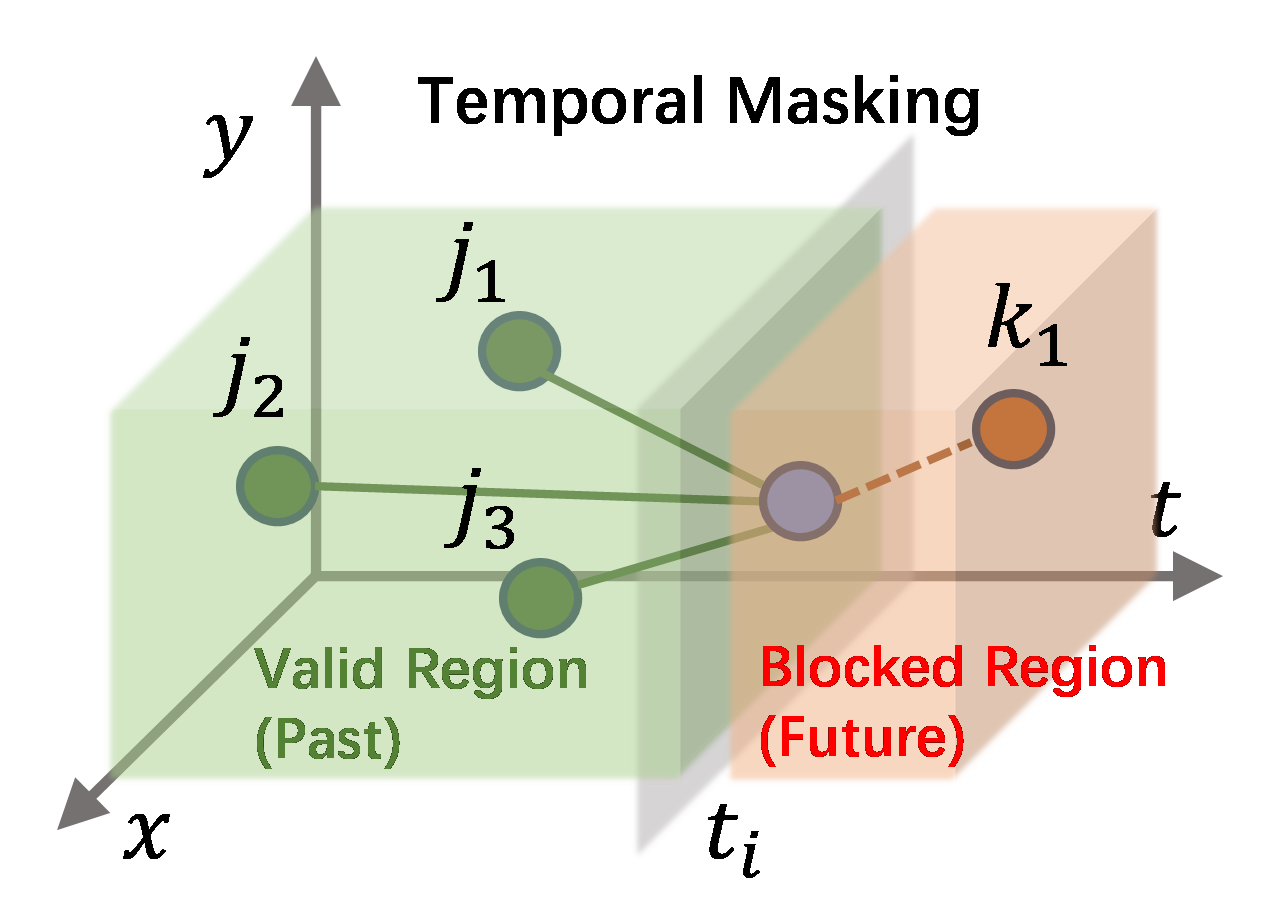}
\vspace{-10pt}
\caption{Causal neighbor masks future events, 
ensuring each point only aggregates few 
information from its past.
}
\label{masking}
\vspace{-15pt}
\end{figure}

\subsection{Spatial Encoder}

For many traditional window-based approaches, such as frame accumulation, point clouds processing, or voxelization, the sampling window latency \(\mathcal{L}_{s}\) is significantly larger than the event latency \(\mathcal{L}_{e}\), making the effective processing rate nearly equivalent to that of conventional frame-based cameras (around 20\,FPS). To address this limitation, we propose a lightweight pointwise spatial modeling strategy.

An important consideration for real-time performance is how to balance inference latency and sampling latency—a challenge that has hindered many pointwise modeling approaches (e.g., GNNs) from achieving real-time efficiency. A detailed discussion is provided in Subsection~D.

For each event \(e_i\) in the raw stream \(\mathcal{E}\), we compute interactions only with a small fixed number \(k\) nearest of causal neighbors, represents as \(\mathcal{N}_i^{(k)} \), illustrated in Fig.~\ref{masking}.

Specifically, \(\mathcal{N}_i^{(k)} \) is defined as:
\begin{equation}
    \mathcal{N}_i^{(k)} = 
    \big\{ e_j \in \mathcal{E} \;\big|\; j < i,\; \| (x_i,y_i,t_i) - (x_j,y_j,t_j) \| \leq \mathcal{O} \big\}, 
\label{eq:neighbor}
\end{equation}
where \(\mathcal{O}\) defines the spatio-temporal neighborhood radius based on weighted distance.
Most \(k\) neighbors are selected within the neighborhood range \(\mathcal{O}\), If fewer than 
\(k\) valid neighbors exist, the missing entries are padded with an invalid-neighbor mask. 
\begin{equation}
m_{ij} =
\begin{cases}
1, & \text{if } e_j \in \mathcal{N}_i^{(k)}, \\
0, & \text{otherwise}.
\end{cases}
\label{eq:mask}
\end{equation}

In the neighborhood range \(\mathcal{O}\), we separately process the center event \(e_i\) and its neighboring events \(\mathcal{N}_i^{(k)} \) to construct enriched local representations.

For \(e_i\), we first estimate a local event rate \(\mathcal{R}_i\) and inject this rate cue explicitly; we then concatenate it with \(\mathbf{C}\) and \(p_i\) to form the raw input, followed by a linear projection to obtain a compact center representation.

For a time horizon \(\tau>0\), the local event rate \(\mathcal{R}_i\) is
\begin{equation}
    \mathcal{R}_i \;=\; \frac{ \big|\{\, j \mid t_i-\tau \le t_j < t_i \,\}\big| }{ \tau }.
\label{eq:local-rate}
\end{equation}

The raw feature and the projected center feature are
\begin{equation}
    \mathbf{f}_i^{\text{raw}} \;=\; [\,x_i,\,y_i,\,t_i,\,r_i,\,\boldsymbol{\pi}_i\,] \in \mathbb{R}^{d_{\text{raw}}},
\end{equation}
\begin{equation}
    \tilde{\mathbf{f}}_i \;=\; \mathbf{f}_i^{\text{raw}}\mathbf{W}_c + \mathbf{b}_c, 
    \quad \mathbf{W}_c\!\in\!\mathbb{R}^{d_{\text{raw}}\times D},\; \mathbf{b}_c\!\in\!\mathbb{R}^{D}.
\end{equation}

For \(\mathcal{N}_i^{(k)} \), We rank candidate neighbors by a spatial temporal weighted distance \(\{d | d^{\text{s}}_{ij}, d^{\text{t}}_{ij}\}\) under causality, select the top \(k\), and compute each neighbor's relative offsets \(\Delta \mathbf{d}_{ij}\) and spatio-temporal distance \(d_{ij}\) to the center \(e_i\). These are fed to a two-layer MLP position encoder to produce a positional bias for neighbor features.
\begin{equation}
    d^{\text{s}}_{ij} = \big\| (x_i,y_i) - (x_j,y_j) \big\|_2, \quad d^{\text{t}}_{ij} = \alpha\,|t_i - t_j|
\end{equation}
\begin{equation}
    d_{ij} = \lambda_{s} d^{\text{s}}_{ij} + \lambda_{s} d^{\text{t}}_{ij}
\end{equation}

The relative offsets and position encoder are
\begin{equation}
    \Delta \mathbf{c}_{ij} = [\,x_j-x_i,\, y_j-y_i,\, t_j-t_i\,],
\end{equation}
\begin{equation}
    \mathbf{z}_{ij} = [\,\Delta \mathbf{c}_{ij},\; d_{ij}\,] \in \mathbb{R}^{4},
\end{equation}
\begin{equation}
    \boldsymbol{\phi}_{ij} = \psi(\mathbf{z}_{ij}) = \mathbf{W}_2\, \sigma(\mathbf{W}_1 \mathbf{z}_{ij} + \mathbf{b}_1) + \mathbf{b}_2 \in \mathbb{R}^{D}, 
\end{equation}
where \(\psi(\cdot)\) is a two-layer MLP with ReLU \(\sigma(\cdot)\).
The position-enhanced neighbor feature is: 
\begin{equation}
\mathbf{h}_{ij} \;=\; \tilde{\mathbf{f}}_{j} + \boldsymbol{\phi}_{ij}, \qquad j \in \mathcal{N}_i^{(k)}.
\label{eq:neighbor-feat}
\end{equation}

We use the center feature as the query \((Q)\) and neighbor features as keys/values \((K/V)\) to perform single-query multi-head attention within the local neighborhood. This avoids global \(\mathcal{O}(N^2)\) computation and keeps the per-event cost at approximately \(\mathcal{O}(kD)\), sustaining stable throughput even under high event densities.
\begin{equation}
    \mathbf{Q}_i = \tilde{\mathbf{f}}_i, \quad
    \mathbf{K}_i = [\,\mathbf{h}_{ij}\,]_{j\in \mathcal{N}_i^{(k)}}, \quad
    \mathbf{V}_i = [\,\mathbf{h}_{ij}\,]_{j\in \mathcal{N}_i^{(k)}},
\end{equation}
\begin{equation}
    \boldsymbol{\alpha}_i = \operatorname{softmax}\!\left( \frac{\mathbf{q}_i \mathbf{K}_i^{\top}}{\sqrt{D}} + \mathbf{M}^{\text{pad}}_i \right), \\
\end{equation}
\begin{equation}
    \mathbf{a}_i = \boldsymbol{\alpha}_i \mathbf{V}_i, 
\end{equation}
where \(\mathbf{M}^{\text{pad}}_i\) is the key–padding mask for invalid/padded neighbors.

Finally, the attention output is stabilized with a residual connection and layer normalization to produce robust local spatial features:
\begin{equation}
    \mathbf{z}_i = \tilde{\mathbf{f}}_i + \operatorname{Dropout}(\mathbf{a}_i), 
    \quad \mathbf{y}_i = \operatorname{LN}(\mathbf{z}_i).
\end{equation}

These features are then fed into the subsequent Mamba state-space model for long-range temporal modeling, effectively ``threading'' the localized spatial features across time.

\subsection{Mamba-based Temporal Modeling}

After obtaining the localized spatial sequence $\mathbf{Y}\!\in\!\mathbb{R}^{B\times L\times D}$ from the spatial encoder, we employ a stack of state-space blocks to capture long-range temporal dependencies while supporting both offline parallel training and online streaming inference.

We adopt $M$ stacked Mamba blocks \(\mathcal{M}\) with LayerNorm, Mamba operator, dropout, and residual connection. For block index $\ell=1,\dots,M$:
\begin{align}
\mathbf{H}^{(\ell)} &= \operatorname{LN}\!\big(\mathbf{Y}^{(\ell)}\big), \\
\tilde{\mathbf{H}}^{(\ell)} &= \mathcal{M}\!\big(\mathbf{H}^{(\ell)}\big), \\
\mathbf{Y}^{(\ell+1)} &= \mathbf{Y}^{(\ell)} + \operatorname{Dropout}\!\big(\tilde{\mathbf{H}}^{(\ell)}\big),
\label{eq:mamba-stack}
\end{align}
with $\mathbf{Y}^{(1)}=\mathbf{Y}$ and the final temporal output $\mathbf{T}=\mathbf{Y}^{(M+1)}\!\in\!\mathbb{R}^{B\times L\times D}$.

For low-latency inference, the sequence is processed chunk-by-chunk. Let $\mathbf{X}^{(c)}\!\in\!\mathbb{R}^{B\times L_c\times D}$ be the $c$-th chunk and $\mathbf{s}^{(\ell,c-1)}$ the carried state at block $\ell$:
\begin{equation}
    \mathbf{H}^{(\ell,c)} = \operatorname{LN}\!\big(\mathbf{X}^{(c)}\big),
\end{equation}
\begin{equation}
    \big(\tilde{\mathbf{H}}^{(\ell,c)},\,\mathbf{s}^{(\ell,c)}\big)= \mathcal{M}_{\ell}^{\text{chunk}}\!\big(\mathbf{H}^{(\ell,c)},\,\mathbf{s}^{(\ell,c-1)}\big),
\end{equation}
\begin{equation}
    \mathbf{Y}^{(\ell,c)} = \mathbf{X}^{(c)} + \operatorname{Dropout}\!\big(\tilde{\mathbf{H}}^{(\ell,c)}\big), 
\end{equation}
where each layer maintains its own causal state. This design preserves temporal continuity across chunks with amortized per-event cost close to $\mathcal{O}(D)$.

When historical context is provided, only current-chunk outputs are sent to the temporal stack and classifier, while the historical part is padded with dummy logits to keep label alignment. Formally, for sample $b$ with history length $h_b$:
\begin{align}
\mathbf{T}_b^{\text{curr}} &= \mathbf{T}_b[h_b{:}L,\ :] \in\mathbb{R}^{(L-h_b)\times D}, \\
\hat{\mathbf{Z}}_b &= 
\begin{bmatrix}
\mathbf{0}_{h_b \times C} \\[2pt]
\operatorname{Head}\!\big(\mathbf{T}_b^{\text{curr}}\big)
\end{bmatrix}
\in\mathbb{R}^{L\times C},
\label{eq:history-pad}
\end{align}
where $\mathbf{0}$ are dummy logits (ignored by loss via label masking), $C$ is the number of classes, and \(\operatorname{Head}(\cdot)\) is the classifier head below.

A lightweight MLP head with LayerNorm produces per-event logits:
\begin{equation}
    \mathbf{U} = \operatorname{LN}\!\big(\mathbf{T}\big), \quad  
\mathbf{H} = \operatorname{ReLU}\!\big(\mathbf{U}\mathbf{W}_1 + \mathbf{b}_1\big),
\end{equation}
\begin{equation}
    \mathbf{H} = \operatorname{Dropout}\!\big(\mathbf{H}\big), \quad 
\mathbf{Z} = \mathbf{H}\mathbf{W}_2 + \mathbf{b}_2,
\end{equation}
where \(\mathbf{W}_1\!\in\!\mathbb{R}^{D\times \frac{D}{2}},\ \mathbf{W}_2\!\in\!\mathbb{R}^{\frac{D}{2}\times C}\), and $\mathbf{Z}\!\in\!\mathbb{R}^{B\times L\times C}$ are the class logits aligned with the input events.

\textbf{Chunked training, streaming inference.} During training, we sample contiguous event chunks for efficiency. At inference time, events are processed in streaming, enabling stable throughput and low end-to-end latency even under high event rates.

\subsection{Adaptive-speed inference}

Nevertheless, we observe that in real-world event streams, the event rate exhibits substantial variability, ranging from as low as $10^3$ events/s to over $10^5$ events/s across different scenarios. If all events are processed in a strictly pointwise manner, the sampling window latency can be reduced to below the microsecond scale ($<1\,\mu$s–$1$\,ms), which in turn imposes extremely stringent requirements on inference latency. Although our model already achieves the smallest parameter scale among comparable approaches, its inference latency remains at the millisecond level. To address this imbalance, we propose a variable-rate inference strategy that adaptively modulates the progression speed to optimally balance sampling latency and inference latency.

Let $s_t$ denote the number of events processed per advancement (shared with the chunk/block length), and let $\mathcal{R}_t$ be the instantaneous event rate (events/s). Under a local Poisson-like arrival, the sampling window latency satisfies
\begin{equation}
L_{\mathrm{win}}(t) \approx \frac{s_t}{\mathcal{R}_t}.
\label{eq:win-lat}
\end{equation}
We target a window-latency budget $L_{\mathrm{win}}^\star$ and an inference-latency budget $L_{\mathrm{inf}}^\star$. The per-step inference time is modeled online by an affine fit
\begin{equation}
\widehat{T}_{\mathrm{inf}}(s) \approx a + b\,s,
\label{eq:tinf-fit}
\end{equation}
estimated from streaming measurements. A conservative base step is obtained from the intersection of the two constraints:
\begin{equation}
s_0(t) \;=\; \min\!\Big(\; \mathcal{R}_t \, L_{\mathrm{win}}^\star,\;\max\{\, s \,\big|\, \widehat{T}_{\mathrm{inf}}(s) \le L_{\mathrm{inf}}^\star \}\;\Big).
\label{eq:base-step}
\end{equation}

To robustly track the window-latency budget under rate fluctuations, we adopt a PID controller on the error
\begin{equation}
e_t \;=\; L_{\mathrm{win}}^\star \;-\; \frac{s_t}{\mathcal{R}_t}.
\label{eq:pid-err}
\end{equation}

The step size is updated as \(s_{t+1}\).
\begin{equation}
s_{t+1} \;=\; \operatorname{clip}\!\Big( s_t + \big( K_P\, e_t + K_I \sum_{\tau \le t} e_\tau + K_D (e_t - e_{t-1}) \big)\, \mathcal{R}_t \Big)
\label{eq:pid-update}
\end{equation}

Then fused with the base step $s_0(t)$ by a convex combination:
\begin{equation}
s_{t+1} \;\leftarrow\; \lambda\, s_0(t) + (1-\lambda)\, s_{t+1}, \qquad \lambda \in [0,1].
\label{eq:blend}
\end{equation}

The multiplication by $\mathcal{R}_t$ in Eq.~\eqref{eq:pid-update} normalizes the error into the event-count domain, improving responsiveness.

To preserve spatial efficiency at high rates, the history size is co-adapted with the step size:
\begin{equation}
H_{t+1} \;=\; \operatorname{round}\!\Big( H_{\text{base}} \cdot \frac{s_{\min}}{\max(s_{t+1},1)} \Big),
\label{eq:hist-adapt}
\end{equation}
so that larger steps (faster progression) reduce the neighborhood history used by the spatial encoder.

In practice, the PID-stabilized variable-rate progression sustains stable throughput under severe rate variability by jointly meeting the sampling and inference latency budgets.

\section{Experiments}

\begin{figure*}[!t]
\centering
\includegraphics[width=7in]{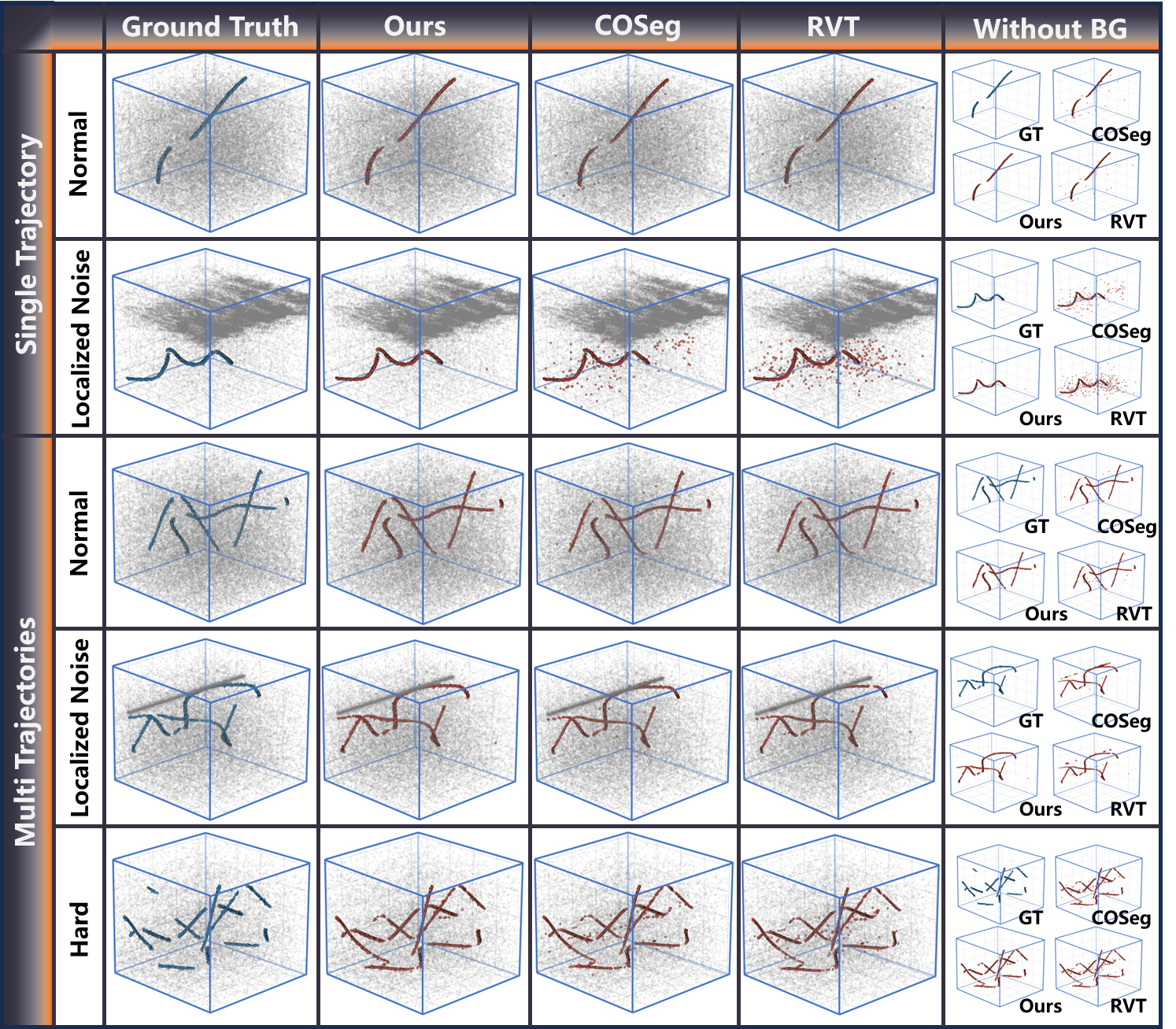}
\vspace{-20pt}
\caption{Qualitative comparison of trajectory segmentation across normal, noisy, and hard cases. Our method yields cleaner and more complete trajectories than COSeg and RVT, even under challenging noise and clutter.}
\label{3}
\vspace{-20pt}
\end{figure*}

\begin{table*}[!t]
\centering
\caption{Quantitative comparison of the proposed method to state-of-the-art methods. The \textbf{bold} and the \underline{underline} represent the best and second-best performance, respectively.}
\vspace{-5pt}
\label{ACC}
\resizebox{\linewidth}{!}{
\begin{tabular}{lcccccccccc}
\hline
\multirow{2}{*}{\textbf{Methods}} & \multirow{2}{*}{\textbf{Year}} & \multirow{2}{*}{\textbf{Event Rep.}} & \multicolumn{2}{c}{\textbf{Window Level}} & \multicolumn{4}{c}{\textbf{Point Level}} & \multirow{2}{*}{\textbf{\#Params.}}  \\
\cline{4-5} \cline{6-9}
& & & $P_d(\%)\uparrow$ & $F_a(10^{-4})\downarrow$ & $P_d(\%)\uparrow$ & $F_a(10^{-4})\downarrow$ & $IoU(\%)\uparrow$ & $ACC(\%)\uparrow$ & \\
\hline
SSD~\cite{liu2016ssd}                  & 2016 & Event Count    & 26.56 & 384.84 & 27.40 & 3616.23 & 33.31 & 37.41 & 25.3M  \\
Faster RCNN~\cite{ren2015faster}       & 2016 & Event Count    & 27.65 & 545.42 & 28.53 & 5093.73 & 48.61 & 38.88 & 41.4M \\
DETR~\cite{zhu2020deformable}          & 2020 & Event Count    & 31.94 & 499.30 & 32.96 & 4663.07 & 39.95 & 44.06 & 39.9M  \\
YOLOV10-S~\cite{wang2024yolov10}       & 2025 & Event Count    & 32.48 & 466.32 & 33.52 & 4355.09 & 42.84 & 43.74 & 7.5M  \\
\hline
EMS-YOLO~\cite{su2023deep}             & 2023 & SNN            & 51.15 & 88.86  & 52.79 & 829.85  & 48.40 & 56.23 & 3.3M  \\
Spike-YOLO~\cite{luo2024integer}       & 2024 & SNN            & 60.18 & 43.80  & 62.10 & 409.02  & 57.84 & 63.22 & 69.0M  \\
GET~\cite{peng2023get}                 & 2023 & Groupy Token   & 61.30 & 36.65  & 63.25 & 342.32  & 53.06 & 64.07 & 18.8M \\
RED~\cite{perot2020learning}           & 2020 & Voxel Grid     & 54.26 & 80.88  & 55.99 & 755.33  & 47.37 & 59.66 & 24.3M  \\
RVT~\cite{gehrig2023recurrent}         & 2023 & Voxel Grid     & 60.91 & 44.03  & 62.86 & 411.23  & 56.87 & 67.31 & 10.0M  \\
SAST~\cite{peng2024scene}              & 2024 & Voxel Grid     & 51.69 & 118.88 & 53.34 & 1110.21 & 45.16 & 52.69 & 18.6M \\
\hline
KPConv~\cite{thomas2019kpconv}         & 2019 & Points         & 69.23 & 12.91  & 71.44 & 120.53  & 63.43 & 75.04 & 50.1M  \\
RandLA-Net~\cite{hu2020randla}         & 2020 & Points         & 71.22 & 5.50   & 73.49 & 51.33   & 66.23 & 77.67 & \underline{1.2M} \\
COSeg~\cite{an2024rethinking}          & 2024 & Points         & 71.99 & 7.28   & 74.28 & 68.02   & 68.30 & 79.82 & 23.4M  \\
EV-SpSegNet~\cite{chen2025event}       & 2025 & Points         & \underline{78.26} & \underline{1.29} & \underline{80.75} & \underline{12.04} & \underline{72.59} & \underline{85.18} & 4.0M  \\
\textbf{Ours}                          & -    & Event Stream   & \textbf{80.57} & \textbf{0.13} & \textbf{91.82} & \textbf{6.44} & \textbf{86.40} & \textbf{92.29} & \textbf{0.21M}  \\
\hline
\end{tabular}
}

\footnotesize{
Our method does not rely on fixed windows; its latency is primarily determined by the adaptive step size. For other window-based approaches, we accumulation a \textbf{50\,ms} window following the official EV-UAV \cite{chen2025event} benchmark. Meanwhile, we additionally probed shorter windows to narrow the latency gap, but found that when the window length was reduced below 20\,ms, these methods failed to produce stable positives (Pd collapsed).
}

\end{table*}


\begin{table}[!t]
\centering
\caption{Latency comparison showing window latency $\mathcal{L}_w$, inference latency $\mathcal{L}_i$, and total latency $\mathcal{L}$ across methods.}
\vspace{-10pt}
\label{tab:Latency}
\resizebox{\linewidth}{!}{
\begin{tabular}{lcccc}
\hline
\multirow{2}{*}{\textbf{Methods}}  & \multicolumn{2}{c}{\textbf{\(\mathcal{L}_{w}\)(ms)}} & \multirow{2}{*}{\textbf{\(\mathcal{L}_{i}\)(ms)}} & \multirow{2}{*}{\textbf{\(\mathcal{L}\)(ms)}} \\
\cline{2-3}
 & Fixed & Dynamic  \\
\hline
SSD~\cite{liu2016ssd}                & 50 & - & 26.85 & 76.85  \\
Faster RCNN~\cite{ren2015faster}     & 50 & - & 33.48 & 83.48  \\
DETR~\cite{zhu2020deformable}        & 50 & - & 22.30 & 72.30  \\
YOLOV10-S~\cite{wang2024yolov10}     & 50 & - & 10.45 & 60.45  \\
\hline
EMS-YOLO~\cite{su2023deep}           & 50 & - & 12.94 & 62.94  \\
Spike-YOLO~\cite{luo2024integer}     & 50 & - & 17.63 & 67.63  \\
GET~\cite{peng2023get}               & 50 & - & 17.79 & 67.79  \\
RED~\cite{perot2020learning}         & 50 & - & 23.54 & 73.54  \\
RVT~\cite{gehrig2023recurrent}       & 50 & - & 24.35 & 74.35  \\
SAST~\cite{peng2024scene}            & 50 & - & 24.86 & 74.86  \\
\hline
KPConv~\cite{thomas2019kpconv}       & 50 & - & 4.55 & 54.55  \\
RandLA-Net~\cite{hu2020randla}       & 50 & - & 3.24 & 53.24 \\
COSeg~\cite{an2024rethinking}        & 50 & - & 2.98 & 52.98  \\
EV-SpSegNet~\cite{chen2025event}     & 50 & - & \textbf{0.79} & \underline{50.79}  \\
\hline
\textbf{Ours} & - & \textbf{\(\sim\)0.57} & \underline{\(\sim\)2.83} & \textbf{3.40}  \\
\hline
\end{tabular}
\vspace{-10pt}
}

\end{table}

\subsection{Dataset and Implementation Details}

We evaluate on the \textbf{EV-UAV} dataset {\cite{chen2025event}}, which contains UAV-mounted event streams with pixel-level foreground/background annotations and widely varying event rates across sequences. 

Our model uses hidden dimension $128$, state dimension $32$, $k{=}16$ causal neighbors within a spatial radius of $0.2$, and dropout $0.1$. We train for 30 epochs with AdamW (lr=$1\times 10^{-4}$, weight decay=$1\times 10^{-5}$), batch size 8, and focal loss ($\alpha{=}0.5$, $\gamma{=}2.0$) with history labels masked. Gradient clipping ($1.0$ and early stopping (patience 5) are applied, and inference is strictly causal with event-rate–adaptive step sizes.

\subsection{Evaluation Metrics}
We adopt rigorous evaluation protocols that assess both strict \emph{point-level} event classification and \emph{window-level} (object-compatible) detection. Unlike frame-based or relaxed metrics, our point-level protocol enforces per-event assessment to faithfully measure temporal precision, while the window-level protocol aggregates within short time bins to compare fairly with box/object detectors, as shown in Table~\ref{ACC}.

\textbf{Segmentation / Point-level Metrics.}
For each file we compute true/false positives and negatives, 
detection probability (recall) $\mathrm{Pd}=\tfrac{TP}{TP+FN}$, false-alarm rate per background event $\mathrm{Fa}=\tfrac{FP}{N_{\text{bg}}}$, positive-class intersection-over-union
\begin{equation}
\text{IoU}_{\text{pos}}=\frac{|TP|}{|TP|+|FP|+|FN|},
\end{equation}
and precision (Prec) is defined as $\mathrm{Prec}=\tfrac{TP}{TP+FP}$.

\textbf{Detection / Window-level (object-compatible) Metrics.}
Each file’s is devided into equally sampling windows, for each window, ground-truth (GT) objects are defined by rasterizing GT-positive events to a $W{\times}H$ grid and extracting 8-connected components. A GT object $o$ (pixel set $\Omega_o$) is counted as detected if the coverage of predicted positives reaches a threshold $\tau_c$:
\begin{equation}
\mathrm{cover}(o;\theta)=\frac{|\hat Y^{(\theta)}\cap \Omega_o|}{|\Omega_o|}\;\ge\;\tau_c .
\end{equation}
Across all files and bins, the \emph{detection probability} is $\mathrm{Pd}=\tfrac{N_{\text{det}}}{N_{\text{obj}}}$, and the \emph{false-alarm density} normalizes the number of false-positive connected components per pixel per frame (time bin):
\begin{equation}
\mathrm{Fa}=\frac{N_{\text{fp-comp}}}{N_{\text{bins}}\cdot W\cdot H}.
\end{equation}

Table~\ref{ACC} summarizes the quantitative comparison with state-of-the-art methods. 
Our approach achieves the highest detection probability and accuracy at both window- and point-level evaluation, while maintaining the lowest false-alarm rate and the smallest parameter scale by a large margin.


\subsection{Latency evaluation}
We evaluate latency by measuring the end-to-end time cost from the arrival of the first event in a chunk until the completion of inference. 

Unlike most existing works that only report the inference latency $\mathcal{L}_i$, we emphasize the overall delay $\mathcal{L}$, which accounts for both the sampling window latency $\mathcal{L}_w$ and the inference latency. 

This distinction is crucial, as event-based systems are deployed in highly dynamic scenarios where even sub-millisecond delays may influence downstream tasks. 

As shown in Table~\ref{tab:Latency}, frame-based methods require a fixed accumulation window (e.g., 50\,ms), leading to overall delays on the order of tens of milliseconds regardless of their inference speed. 

Our variable-rate strategy eliminates the fixed window and adaptively balances $\mathcal{L}_w$ and $\mathcal{L}_i$, achieving $\mathcal{L}_w\!\approx\!0.57$\,ms, $\mathcal{L}_i\!\approx\!2.83$\,ms, and overall $\mathcal{L}\!\approx\!3.40$\,ms—an order of magnitude lower than existing methods.

\begin{figure}[!t]
\centering
\includegraphics[width=3.2in]{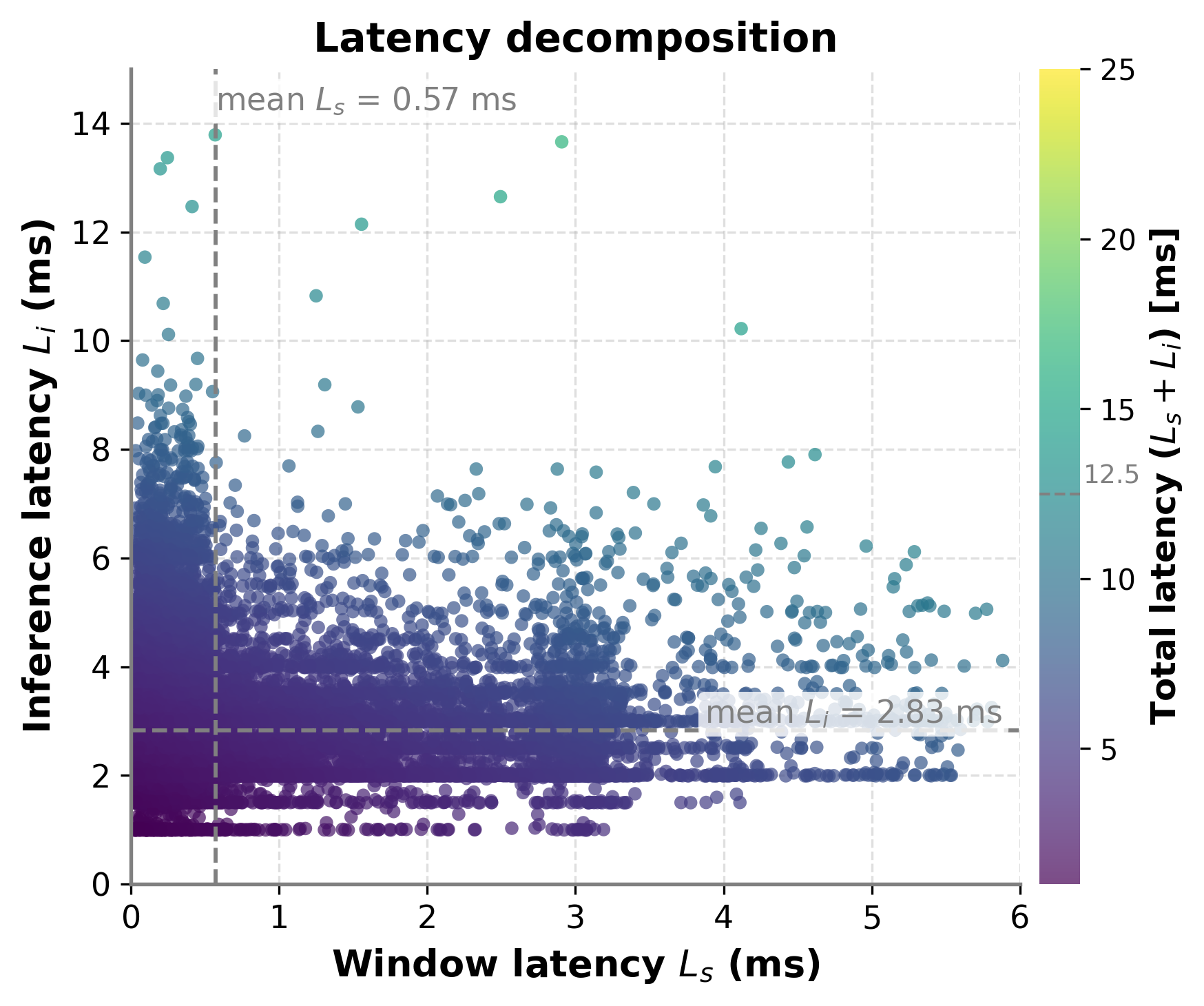}
\vspace{-15pt}
\caption{Latency decomposition of our method, showing mean window latency $L_s=0.57$\,ms and mean inference latency $L_i=2.83$\,ms, resulting in consistently low end-to-end delay.}
\label{3}
\vspace{-15pt}
\end{figure}

\subsection{Ablation studies}
We conducted ablation experiments to analyze the effect of different design choices in our framework, as summarized in Table~\ref{tab:ablation}. 

For spatial encoding, removing the KNN neighborhood reduces the per-step spatial aggregation cost (no neighbor search/attention), thus slightly lowers latency. However, the model loses local spatial consistency cues, leading to a notable Pd/IoU drop. This is expected: KNN attention is the main vehicle to encode fine-grained spatial correlations among nearby events. The grid-based and voxel-based encoders are significantly
worse than our method while maintaining the same order of magnitude of parameters, and have longer inference time.

For temporal modeling, the Mamba module is replaced with a recurrent model (LSTM/GRU). Although there is little difference in detection accuracy, there is a significant delay in streaming compared to Mamba.

We also varied hyperparameters such as the neighbor size $k$ and hidden dimension $d_h$, where $k=16$ provided the best balance between detection probability, false alarms, and model size.

\begin{table}[!t]
\centering
\caption{Component analysis and design choices}
\vspace{-10pt}
\label{tab:ablation}
\begin{tabular}{lcccc}
\toprule
\textbf{Configuration} & \textbf{$P_d(\%)\uparrow$} & $F_a(10^{-4})\downarrow$ & \textbf{Params} & {\textbf{\(\mathcal{L}\)(ms)}}\\
\midrule
Full Model & \underline{91.82} & \underline{6.44} & \underline{0.21M} & \underline{3.40} \\ 
\midrule
\multicolumn{5}{l}{\textit{Spatial Encoding Variants}} \\
w/o Spatial K-NN & 73.47 & 9.96 & 0.21M & \textbf{2.47}\\
grid & 72.27 & 13.24 & 0.21M & 16.65\\
voxel & 71.83 & 9.82 & 0.23M & 21.73\\
\midrule
\multicolumn{5}{l}{\textit{Temporal Modeling}} \\
LSTM instead of Mamba & 91.71 & \textbf{6.24} & \textbf{0.18M} & 56.45\\
GRU instead of Mamba & \textbf{92.48} & 6.45 & 0.22M & 55.29\\
\midrule
\multicolumn{5}{l}{\textit{Hyperparameter Sensitivity}} \\
$k=8$ neighbors & 90.96 & 6.53 & 0.21M & 3.37\\
$k=32$ neighbors & 91.71 & 6.49 & 0.21M & 3.51\\
\bottomrule
\end{tabular}
\vspace{-15pt}
\end{table}

\section{Conclusion}
We presented the Variable-Rate Spatial Event Mamba, a lightweight framework that directly processes asynchronous event streams without windowed representations. By combining a causal neighborhood encoder with Mamba state-space modeling and a rate-adaptive controller, our method achieves state-of-the-art performance with minimal parameters and latency. These results underscore the potential of stream-causal, rate-adaptive modeling for real-time event-based perception on UAVs and resource-constrained platforms.

\section*{Acknowledgments}  
The authors acknowledge the use of large language models (e.g., Claude by Anthropic, ChatGPT by OpenAI) for assisting in prototype code generation and improving the clarity of the manuscript. All technical contributions and interpretations remain the responsibility of the authors.

\bibliographystyle{IEEEtran}
\bibliography{refs}

\end{document}